\documentclass{article}
\usepackage{ijcai21}

\usepackage{times}
\usepackage{soul}
\usepackage{url}
\usepackage[hidelinks]{hyperref}
\usepackage[utf8]{inputenc}
\usepackage[small]{caption}
\usepackage{graphicx}
\usepackage{amsmath}
\usepackage{amsthm}
\usepackage{booktabs}
\usepackage{algorithm}
\usepackage{algorithmic}
\usepackage{amsmath,amssymb}
\usepackage{mathtools}
\usepackage{bm}

\usepackage[subrefformat=parens]{subcaption}

\title{RLTutor: Reinforcement Learning Based Adaptive Tutoring System\\ by Modeling Virtual Student with Fewer Interactions}

\author{
Yoshiki Kubotani$^1$\footnote{Contact Author}\and
Yoshihiro Fukuhara$^1$\and
Shigeo Morishima$^2$
\affiliations
$^1$Waseda University\\
$^2$Waseda Research Institute for Science and Engineering
\emails
yoshikikubotani@akane.waseda.jp,
f\_yoshi@ruri.waseda.jp,
shigeo@waseda.jp
}

\begin{document}

\maketitle

\begin{abstract}
  A major challenge in the field of education is providing review schedules that present learned items at appropriate intervals to each student so that memory is retained over time.
  In recent years, attempts have been made to formulate item reviews as sequential decision-making problems to realize adaptive instruction based on the knowledge state of students.
  It has been reported previously that reinforcement learning can help realize mathematical models of students learning strategies to maintain a high memory rate.
  However, optimization using reinforcement learning requires a large number of interactions, and thus it cannot be applied directly to actual students.
  In this study, we propose a framework for optimizing teaching strategies by constructing a virtual model of the student while minimizing the interaction with the actual teaching target.
  In addition, we conducted an experiment considering actual instructions using the mathematical model and confirmed that the model performance is comparable to that of conventional teaching methods.
  Our framework can directly substitute mathematical models used in experiments with human students, and our results can serve as a buffer between theoretical instructional optimization and practical applications in e-learning systems.
\end{abstract}

\section{Introduction}
  The demand for online education is increasing as schools around the world are forced to close due to the COVID-19 pandemic.
  E-learning systems that support self-study are gaining rapid popularity.
  While e-learning systems are advantageous in that students can study from anywhere without gathering, they also have the disadvantage of making it difficult to provide an individualized curriculum, owing to the lack of communication between teachers and students~\cite{Zounek2013heads}.

  Instructions suited to individual learning tendencies and memory characteristics have been investigated recently~\cite{pavlik2008using,khajah2014maximizing}. Flashcard-style questions, where the answer to a question is uniquely determined, have attracted significant research attention.
  Knowledge tracing~\cite{corbett1994knowledge} aims to estimate the knowledge state of students based on their learning history and uses it for instruction. In this context, it has been reported that the success or failure of future students' answers can be accurately estimated using psychological findings~\cite{lindsey2014improving,choffin2019das3h} and deep neural networks (DNNs)~\cite{piech2015deep}.

  Furthermore, adaptive instructional acquisition has also been formulated as a continuous decision problem to optimize the instructional method~\cite{rafferty2016faster,whitehill2017approximately,reddy2017accelerating,upadhyay2018deep}.
  Reddy et al.~\shortcite{reddy2017accelerating} considered the optimization of student instruction as an interaction between the environment (students) and the agent (teacher), and they attempted to optimize the instruction using reinforcement learning.
  Although their results outperform existing heuristic teaching methods for mathematically modelled students, there is a practical problem in that it cannot be applied directly to actual students. This is because of the extremely large number of interactions required for optimization.

  In this study, we propose a framework that can optimize the teaching method while reducing the number of interactions between the environment (modeled students) and the agent (teacher) using a pretrained mathematical model of the students.
  Our contributions are summarized as follows.
  \begin{itemize}
      \item We pretrained a mathematical model that imitates students using mass data, and realized adaptive instruction using existing reinforcement learning with a smaller number of interactions.
      \item We conducted an evaluation experiment of the proposed framework in a more practical setting, and showed that it can achieve comparable performance to existing methods with fewer interactions.
      \item We highlighted the need to reconsider the functional form of the loss function for the modeled students to realize more adaptive instruction.
  \end{itemize}
  %

\section{Related Work}
  \subsection{Knowledge Tracing}
  \label{knowledge_tracing}
  Knowledge tracing (KT)~\cite{corbett1994knowledge} is a task used to estimate the time-varying knowledge state of learners from the corresponding learning histories.
  Various methods have been proposed for KT, including those using Bayesian estimation~\cite{yudelson2013individualized}, and DNN-based methods~\cite{piech2015deep,pandey2019self}.

  In this study, we focus on the item response theory (IRT)~\cite{frederic1952irt}.
  IRT is aimed at evaluating tests that are independent of individual abilities. The following is the simplest logistic model:
  \begin{equation}
    \mathbb{P}  \lparen O_{i,j}= 1 \rparen
    =
    \sigma \lparen \alpha_i - \delta_j \rparen. \label{eq:irt}
  \end{equation}
  Here, $\alpha_i$ is the ability of the learner $i$, $\delta_j$ is the difficulty of the question $j$, and $O_{i,j}$ is a random variable representing the correctness $o$ of learner $i$'s answer to question $j$ in binary form.
  $\sigma\lparen\cdot\rparen$ represents the sigmoid function.
  While ordinary IRT is a static model with no time variation, IRT-based KT attempts to realize KT by incorporating learning history into Equation \eqref{eq:irt}~\cite{cen2006learning,pavlik2009performance,Vie_Kashima_2019}.
  Lindsey et al.~\shortcite{lindsey2014improving} proposed the DASH model, which uses psychological knowledge to design a history parameter.
  Since DASH does not consider the case where multiple pieces of knowledge are associated with a single item, Choffin et al.~\shortcite{choffin2019das3h} extended DASH to consider correlations between items, and the model is called DAS3H:
  \begin{align}
    &\mathbb{P}  \lparen O_{i,j,l} = 1 \rparen \notag\\
    &=
    \sigma\Bigl\lparen
                \alpha_i - \delta_j
                +
                \displaystyle \sum_{k \in \mathrm{KC}_j}{\beta_k}
                +
                h_{\theta, \phi} \lparen t_{i,j,1:l}, o_{i,j,1:l-1} \rparen
    \Bigl\rparen, \label{eq:das3h_main} \\
    &h_{\theta, \phi} \lparen t_{i,j,1:l}, o_{i,j,1:l-1} \rparen \notag\\
    &=
    \sum_{k \in \mathrm{KC}_j}
    \sum_{w=0}^{W-1}{
        \theta_{k,w} \ln \lparen 1+c_{i,k,w} \rparen
        +
        \phi_{k, w} \ln \lparen 1+n_{i,k,w} \rparen.
    } \label{eq:das3h_h}
  \end{align}
  Equation \eqref{eq:das3h_main} contains two additional terms from Equation \eqref{eq:irt}: the proficiency $\beta_k$ of the knowledge components (KC) associated with item $j$, and $h_{\theta, \phi}$ defined in Equation \eqref{eq:das3h_h}.
  $h_{\theta, \phi}$ represents students' learning history, $n_{i,k,w}$ refers to the number of times learner $i$ attempts to answer skill $k$, and $c_{i,k,w}$ refers to the number of correct answers out of the trials, both counted in each time window $\tau_w$.
  Time window $\tau_w$ is a parameter that originates from the field of psychology~\cite{RoveeCollier1995TimeWI}, and represents the time scale of loss of memory.
  By dividing the counts by each discrete time scale satisfying $\tau_w < \tau_{w+1}$, the memory rate can be estimated by taking into account the temporal distribution of the learning history~\cite{lindsey2014improving}.

  \subsection{Adaptive Instruction}
  A mainstream approach to adaptive instruction is the optimization of review intervals.
  The effects of repetitive learning on memory consolidation have been discussed in the field of psychology~\cite{ebbinghaus1885gedachtnis}, and various studies have experimentally confirmed that gradually increasing the repetition interval is effective for memory retention~\cite{leitner1974so,wickelgren1974single,landauer1978optimum,wixted2007wickelgren,cepeda2008spacing}.

  Previously, the repetition interval was determined algorithmically when the item was presented~\cite{khajah2014maximizing}.
  However, in recent years, there have been some attempts to obtain more personalized instruction by treating such instruction as a sequential decision problem~\cite{rafferty2016faster,whitehill2017approximately,reddy2017accelerating,upadhyay2018deep}.
  Rafferty et al.~\shortcite{rafferty2016faster} formulated student instruction as a partially observed Markov decision process (POMDP) and attempted to optimize instruction for real students through planning for multiple modelled students.
  Based on their formulation, Reddy et al.~\cite{reddy2017accelerating} have also attempted to optimize instructional strategies using trust region policy optimization (TRPO)~\cite{schulman2015trust}, a method of policy-based reinforcement learning.
  However, optimization by reinforcement learning requires a large number of interactions, which makes it inapplicable to real-life scenarios.

\section{Proposed Framework}
  To address the issue of large numbers of interactions, we formulate a framework for acquiring adaptive instruction with fewer contacts.
  In this section, we consider student teaching as a POMDP and formulate a framework for acquiring adaptive teaching with a small number of interactions.

\begin{figure}[tb]
    \centering
    \includegraphics[width=70mm,bb=15 10 340 270]{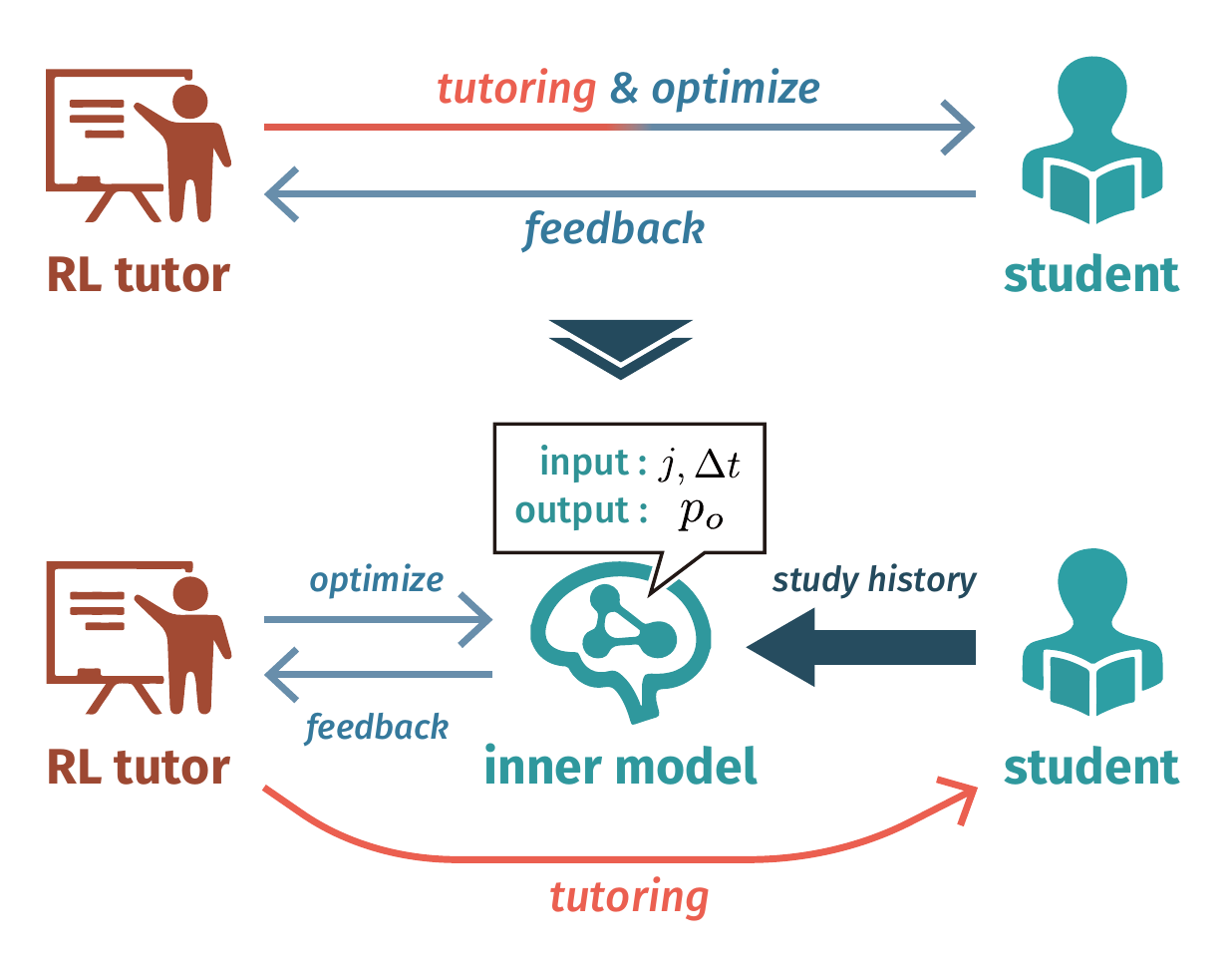}
    \caption{Illustration of the difference between the usual reinforcement learning setting and the proposed method. The proposed method updates the instructional strategies not by interacting with the students directly, but by interacting with the KT model estimated from the student's study history.}
    \label{fig:proposed_framework}
  \end{figure}

  The proposed method has two main structures: a memory model that captures the knowledge state of the student (inner model) and a teaching model that acquires the optimal instructional strategy through reinforcement learning (RLTutor).
  As shown in Figure~\ref{fig:proposed_framework}, RLTutor optimizes its strategy indirectly through interaction with the inner model, rather than with the actual student.
  In the following sections, we first describe the detailed design of the inner model and RLTutor and describe the working principle of our framework.
  \begin{figure*}
    \centering
    \includegraphics[width=128mm, bb=84 5 500 180]{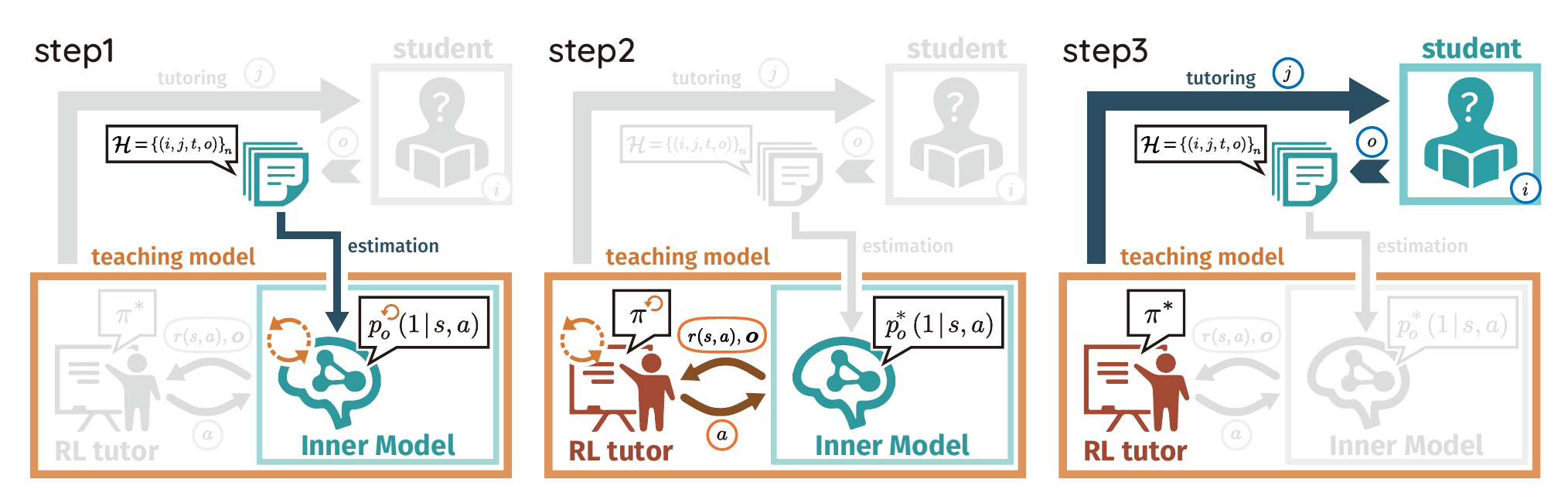}
    \captionof{figure}{Graphical representation of each step of the proposed framework. First, the Inner Model is updated from the learning history of the student. Next, RLTutor optimizes the teaching policies for the updated Inner Model. Finally, RLTutor presents items to the target based on the optimized policy. Before the start of the experiment, the Inner Model is pretrained using the educational dataset, and then updated by minimizing the loss function in Equation~\eqref{eq:loss_fn} using all the historical data obtained up to that point.}
    \label{fig:architecture_flow}
  \end{figure*}

  \subsection{Inner Model}
  Inner Model is a virtual student model constructed from the learning history of the instructional target using the KT method described in Section~\ref{knowledge_tracing}.
  Specifically, given a learning history $\mathcal{H}=\{i,j,t,o\}_n$ with the same notation as in Section~\ref{knowledge_tracing}, the model returns the probability of the correct answer $p_o$ for the next given question.
  In this study, we adopted the DAS3H model expressed in Equations \eqref{eq:das3h_main} and \eqref{eq:das3h_h} as the KT method.
  However, because this model uses a discrete time window to account for past learning, the estimated memory probability may vary from $1$, even if the same item is presented consecutively (e.g., without an interval).
  However, the human brain is capable of remembering meaningless symbols for a few seconds to a minute~\cite{miller1994magical}.
  To capture this sensory memory, we adopted the functional form proposed by Wicklegren~\shortcite{wickelgren1974single} and used:
  \begin{equation}
    p_o \left\lparen O = 1 \mid s,a \right\rparen
    \coloneqq
    \lparen
        1-p_D
    \rparen
    \lparen
        1+h \cdot \Delta t
    \rparen^{-f}
    +
    p_D \label{eq:inner_model}
  \end{equation}
  as the equation for the internal model with $p_\mathrm{D}$ Choffin et al.'s DAS3H model.
  In this study, $h$ and $f$ in Equation~\eqref{eq:inner_model} are regarded as constants, and set to $h=0.3,f=0.7$. This is because individual differences in sensory memory are much smaller than those in long-term memory.

  \subsection{RLTutor}
  RLTutor is an agent that learns an optimal strategy $\pi^*$ using reinforcement learning based on the responses from the inner model, and uses the obtained strategy to provide suboptimal instruction to actual students.

  As for the POMDP formulation, we assume that the true state $\mathcal{S}=\mathbb{R}^{I \times J \times K \times (W \times K)^2} \times {\mathbb{N}_0}^{(I \times K \times W)^2}$ of the student is not available, and only the correct and incorrect responses of the student can be obtained as an observation $\mathcal{O}=\{0,1\}$.
  We also set the immediate reward $r(s,a)$ as the average of the logarithmic memory retention for all problems at that time:
  \begin{equation}
    r(s,a)
    =
    \displaystyle \frac{1}{J} \sum_{j=1}^J
        \log p_o \left\lparen O_j = 1 \mid s,a \right\rparen.
    \label{eq:reward}\\
  \end{equation}
  For the above partial observation problem, we applied the proximal policy optimization (PPO)~\cite{schulman2017proximal} method, which is a derivation of the TRPO.
  The policy function $\pi(a \vert s)$ is comprised of a neural network, including a gated recurrent unit (GRU) with 512 hidden layers.

  Following Reddy et al.~\shortcite{reddy2017accelerating} we defined the input to the network as:
  \begin{align}
    \bm{o}
    &=
    \begin{pmatrix}
        \bm{j}_\mathrm{e} \\
        \bm{t}_\mathrm{e} \\
        o
    \end{pmatrix}
    \in
    \mathbb{R}^{\log{2J} + 2 + 1},
    \label{eq:embedding_input}
  \end{align}
  where $\bm{j}_\mathrm{e}$ is a vector embedded with the presented item number $j$ following Equation~\eqref{eq:vec1} --~\eqref{eq:item_embedding}, $\bm{t}_\mathrm{e}$ is a vector embedded with the time elapsed since the previous interaction $\Delta t$ following Equation~\eqref{eq:time_embedding}, and $o$ is the student's response:
  \begin{align}
    \bm{v}_1
    &=
    \text{onehot}(j)
    \in
    \mathbb{R}^{J}, \label{eq:vec1}\\
    \bm{v}_2
    &=
    \begin{pmatrix}
        \bm{v}_1 \cdot (1-o) \\
        \bm{v}_1 \cdot o
    \end{pmatrix}
    \in
    \mathbb{R}^{2J}, \label{eq:vec2}\\
    \bm{j}_\mathrm{e}
    &=
    g(\bm{v}_2)
    \in
    \mathbb{R}^{\log{2J}}, \label{eq:item_embedding}\\
    \bm{j}_\mathrm{t}
    &=
    \begin{pmatrix}
        \log{\Delta t} \cdot (1-o) \\
        \log{\Delta t} \cdot o
    \end{pmatrix}
    \in
    \mathbb{R}^{2}. \label{eq:time_embedding}
  \end{align}
  Note that $g$ is a random projection function ($\mathbb{R}^{2J} \rightarrow \mathbb{R}^{\log{2J}}$), compressing the sparse one-hot vector to a lower dimension without loss of information~\cite{piech2015deep}.

  \subsection{Working Principle}
  Based on the two components described above, the proposed framework operates by repeating the following three steps (Figure~\ref{fig:architecture_flow}).
  \begin{enumerate}
    \item Estimation: The inner model is updated based on students' learning history. \label{step1}
    \item Optimization: RLTutor optimizes the strategy for the updated inner model. \label{step2}
    \item Instruction: RLTutor poses an item to the student using the current strategy. \label{step3}
  \end{enumerate}

  Because it is difficult to train the inner model from scratch, we pretrained it using EdNet, which is the largest existing educational dataset~\cite{choi2020ednet}. We trained DAS3H on this dataset and determined the initial values of the Inner Model's weights from the distribution of the estimated parameters. For details, please see the code\footnote{The code will soon be available on \url{https://github.com/YoshikiKubotani/rltutor}.} to be released soon. In addition, in step~\ref{step1}, we used the following loss function to prevent excessive deviation from the pre-trained weights:
  \begin{equation}
    \mathcal{L}
    =
    \mathcal{L}^{\mathrm{MSE}}
    +
    \displaystyle \sum_{m}
        c_m \{
            \lparen 1-\lambda \rparen \mathcal{L}_m^{\mathrm{DIST}}
            +
            \lambda \mathcal{L}_m^{\mathrm{FIX}}
        \}. \label{eq:loss_fn}\\
  \end{equation}
  Here, $\mathcal{L}^{\mathrm{MSE}}$ is introduced to make the output $p_o$ of the model closer to the student's response $o$ obtained as the study history.
  $\mathcal{L}_m^{\mathrm{DIST}}$ and $\mathcal{L}_m^{\mathrm{FIX}}$ are constraint terms that prevent each parameter of the model from deviating from the distribution obtained in the prior training and from the value of the previous parameter, respectively.
  These can be expressed as:
  \begin{align}
    \mathcal{L}^{\mathrm{MSE}}
    &\coloneqq
    \displaystyle \sum_n \bigl\lparen
        o - p_o
    \bigr\rparen^2, \label{eq:mse_loss}\\
    \mathcal{L}_m^{\mathrm{DIST}}
    &\coloneqq
    \displaystyle \sum_n \left\lparen
        1 - \frac{
            f_m \lparen \rho_m \rparen
        }{
            f_m \lparen \mu_m \rparen
    }\right\rparen ,
    \label{eq:gaussian_loss}\\
    \mathcal{L}_m^{\mathrm{FIX}}
    &\coloneqq
    \displaystyle \sum_n \vert
        \rho_m - \grave{\rho}_m
    \vert,
    \label{eq:l1_fix_loss}
  \end{align}
  where $f_m(x)$ is the normal distribution with mean $\mu_m$ and variance $\sigma_m$, both of which are obtained during pretraining.
  $\rho_m$ is a generalized representation of each parameter of the DAS3H model($\rho_1=\alpha, \rho_2=\delta, \rho_3=\beta, \rho_4=\theta, \rho_5=\phi$).

  In this manner, we attempted to obtain adaptive instruction while reducing the number of interactions with the teaching target.

\section{Experiment}
  \begin{figure}[tb]
    \centering
    \includegraphics[width=110mm,bb=1 1 340 140]{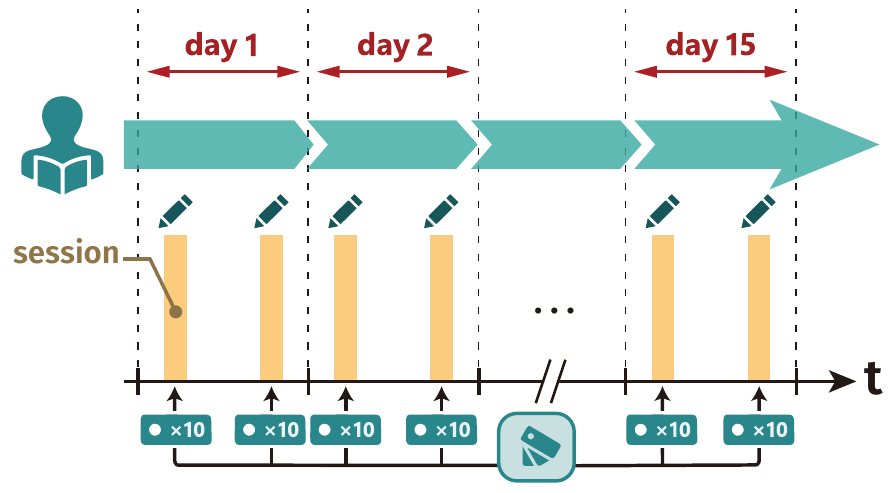}
    \caption{Experimental flow. Students who are substituted in the mathematical model are required to study a total of 30 items through 15 days. There are two intensive learning periods (called ``sessions'') per day, and the teacher agents present 10 items for each session.}
    \label{fig:experiment}
  \end{figure}
  \begin{figure}[bt]
    \centering
    \includegraphics[width=5mm,bb=305 250 340 630]{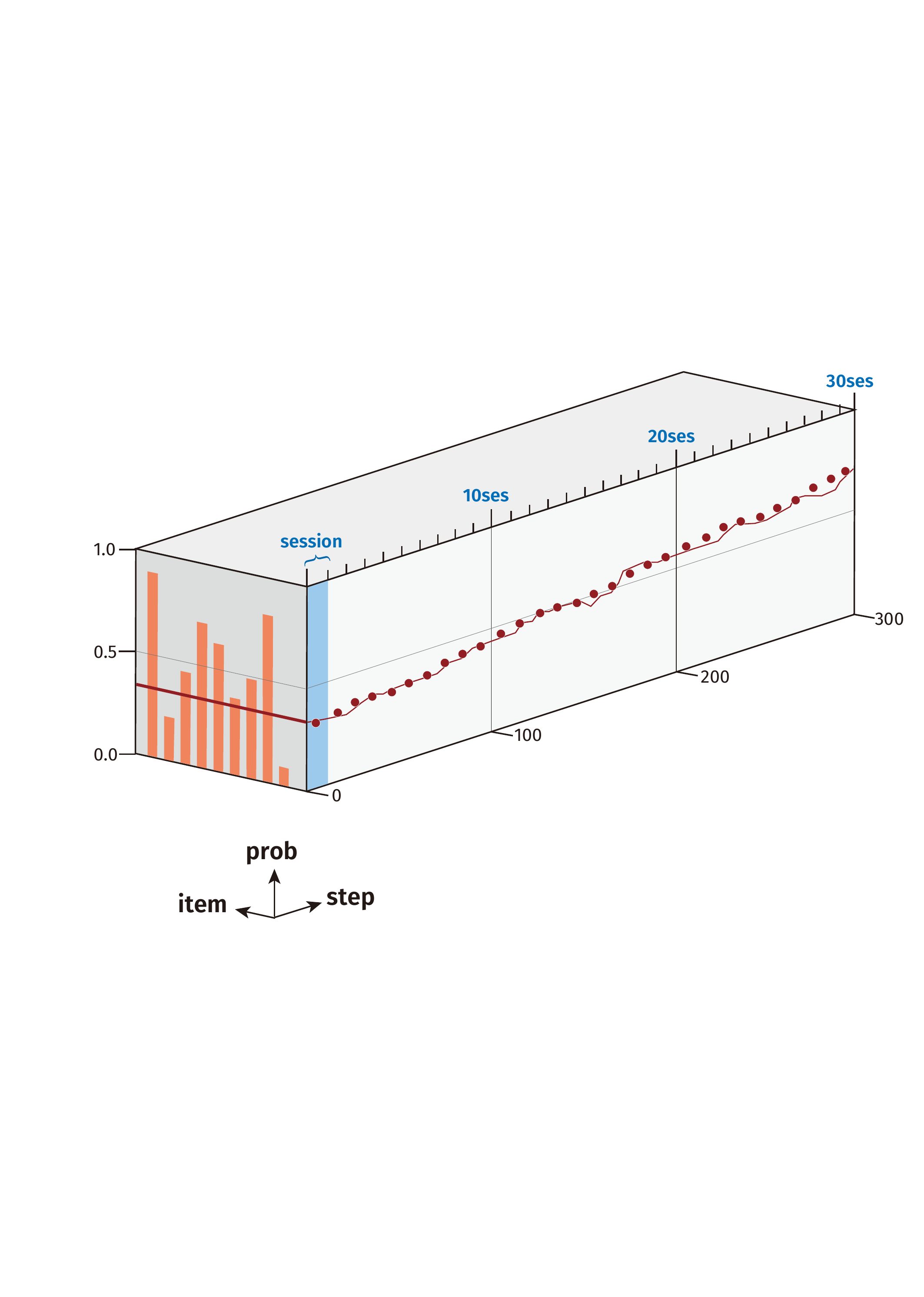}
    \caption{Diagrammatic representation of the plotting mechanism. The mathematical model used on behalf of the students can calculate the memory rate for each item at each step. First, an average over all items is evaluated. Because the responses to the items have been recorded 300 times (30 sessions with 10 items in each session), we then plot the mean values over each session.}
    \label{fig:plot_system}
  \end{figure}
  \begin{figure}[bt]
    \begin{minipage}[b]{\linewidth}
      \centering
      \includegraphics[width=65mm,bb=47 5 340 270]{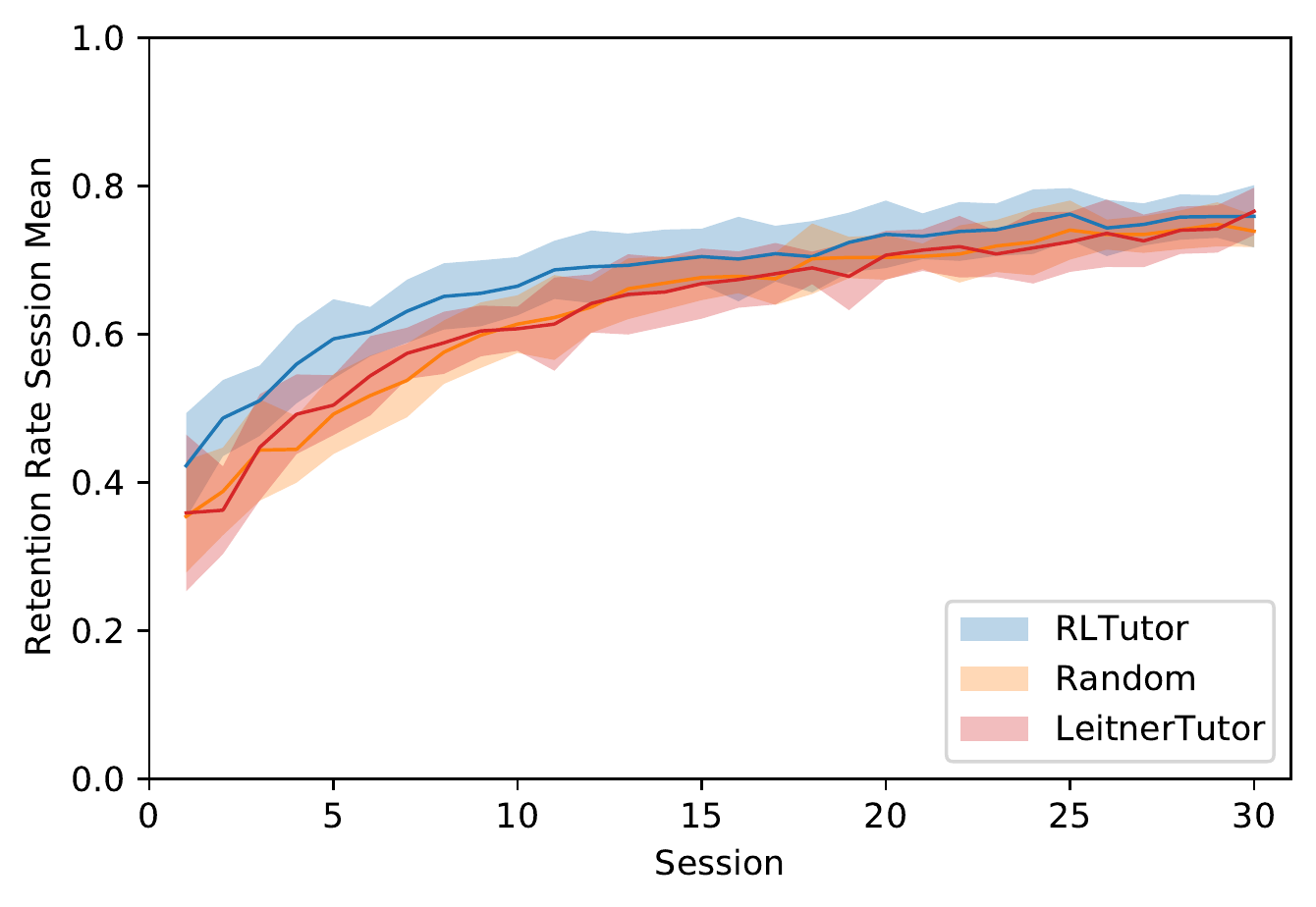}
      \subcaption{VS LeitnerTutor}\label{fig:result_vs_leit}
    \end{minipage}\\
    \begin{minipage}[b]{\linewidth}
      \centering
      \includegraphics[width=65mm,bb=47 5 340 270]{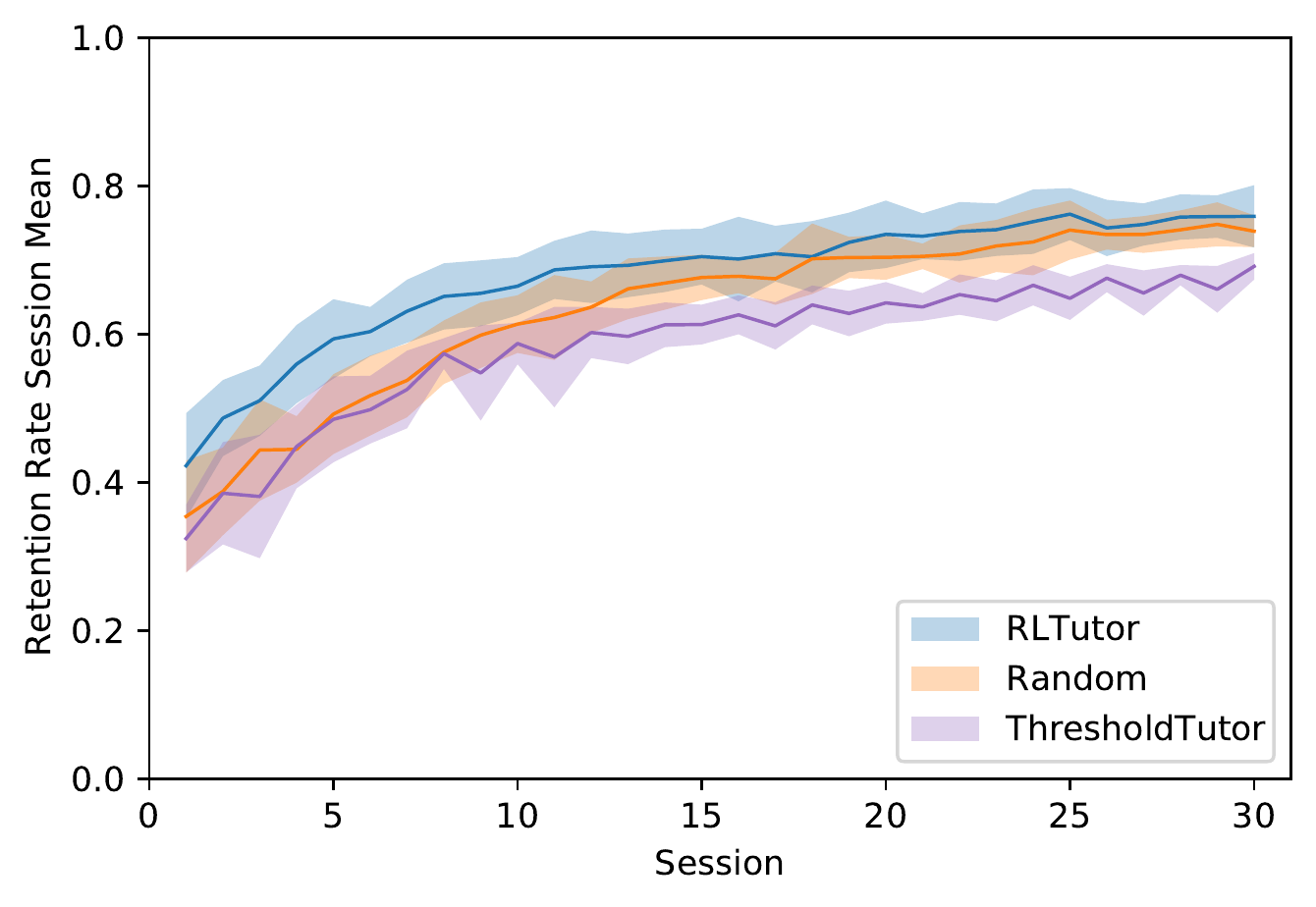}
      \subcaption{VS ThresholdTutor}\label{fig:result_vs_th}
    \end{minipage}
    \caption{Plot of experimental results. We compare LeitnerTutor in Figure~\ref{fig:result_vs_leit} and ThresholdTutor in Figure~\ref{fig:result_vs_th}, in addition to RandomTutor and RLTutor.}\label{fig:result}
  \end{figure}
  To verify the effectiveness of the proposed framework, we conducted a comparison experiment with other instructional methods: RandomTutor, LeitnerTutor, ThresholdTutor, and RLTutor.
  RandomTutor is a teacher that presents problems completely randomly.
  LeitnerTutor is a teacher that presents problems according to the Leitner system, which is a typical interval iteration method.
  ThresholdTutor is a teacher that presents problems based on the student's memory rate closest to a certain threshold value.
  Since user testing with actual students is costly, the mathematical model (DAS3H model) was used for the students taught in this experiment.

  The experimental setup was such that the learner had to study 30 items in 15 days.
  To make the setting more realistic, we set up two sessions per day to study the items intensively, each session presenting 10 items.
  Figure~\ref{fig:experiment} illustrates the outline of the experiment.

  The Inner Model was updated for each session using a batch of all the previous historical data, and it was trained for 10 epochs.
  Note that the coefficients $c_m$ of the constraint terms of the loss function in Equation~\eqref{eq:inner_model} were all set to one.
  For PPO training, mini-batches of size 200 were created every 4000 steps, and 20 agents were trained in parallel for 10 epochs.
  For the hyperparameters, we set the clipping parameter to 0.2, the value function coefficient to 0.5, the entropy coefficient to 0.01, the generalized advantage estimator (GAE) parameter to 0.95, and the reward discount rate to 0.85. The initial learning rate for all parameters was 0.5.
  In both cases, we used the Adam optimizer with an initial learning rate of 0.0001, which was gradually reduced to zero every two sessions.

  The results of the experiments are shown in Figures~\ref{fig:result_vs_leit} and ~\ref{fig:result_vs_th}.
  Figure~\ref{fig:result_vs_leit} shows the results of instruction by RLTutor and LeitnerTutor, and Figure~\ref{fig:result_vs_th} shows the results of instruction by RLTutor and ThresholdTutor.
  The experiments were conducted five times with different seed values, and we plotted the average of the students' memory rates over items and sessions.
  The solid lines represent the mean values, and the colored bands represent the standard deviations.
  To aid comprehension, we illustrate how the item-session average is calculated in Figure~\ref{fig:plot_system}.
  Figure~\ref{fig:result} shows that retention was strengthened in both methods of instruction.
  The figure shows that the instruction by the proposed method successfully maintains a higher memory rate than the other methods.
  In addition, the instruction by ThresholdTutor did not maintain the memory rate of the students as well as other methods.
  However, as the sessions continued, the performance of our method decreased.
  The reasons for these results will be discussed in the Discussion section.

\section{Discussion}
  \subsection{Why are there not much differences?}
  There are three possible reasons for why all the methods improved the students' memory rate with approximately similar results: the experimental setup, loss function of the Inner Model, and gap between the model and the actual students.
  \subsubsection{Experimental Setup}
  The first reason is the experimental setup.
  In this experiment, we set a relatively long learning period (15 days) for a small number of questions (30 questions). Therefore, all questions would have been learned sufficiently, regardless of how they were presented. This leads to similar final memory rates for all methods.

  \subsubsection{Inner Model's Loss Function}
  The next reason is the invariance of the Inner Model.
  As shown in Equation~\eqref{eq:inner_model}, a constraint term was added to the loss function of the Inner Model to avoid any deviation from the pre-learned weights and from the weights learned in the previous sessions.
  It is possible that these constraints are significantly strong, and that the Inner Model was not ``fine-tuned'' as the experiment progressed.

  \subsubsection{Gap between Model and the Actual Students}
  The last possible reason is the accuracy of the model.
  In this experiment, we presented the mathmatical model instead of the actual students.
  The gaps between the model and the actual students may have affected the results.
  In fact, Figure~\ref{fig:result_vs_leit} shows that the instruction by the Leitner system, which is conventionally considered to be effective for memory retention, produced results comparable to random instruction.

  \subsection{Why ThresholdTutor does not work?}
  ThresholdTutor is a method that poses the next item as the one whose memory rate is closest to the threshold value.
  It has been reported that ThresholdTutor performs better than other methods when applied to mathematical models because it can directly refer to the memory rate, which cannot be used normally~\cite{khajah2014maximizing}.
  However, as can be seen from Figure~\ref{fig:result_vs_th}, the results of this experiment are less than that of RandomTutor.
  This is likely due to the number of items to be tackled per session and the characteristics of the models used as students.

  In this experiment, we used the DAS3H model instead of the students. However, depending on the values of $\theta$ and $\phi$, this model could forget after some time, even after repeated learning.
  Therefore, for some items, the recall rate drops to values near the threshold after a half-day session interval, even if the rate is already high.
  In addition, because this experiment was set up to solve 10 items per session, it is considered that ``moderately memorized but easily forgotten items'' were repeatedly presented at the end of the experiment.

\section{Conclusion}
  In this paper, we proposed a practical framework for providing optimal instruction using reinforcement learning based on the estimated knowledge state of the learner. Our framework is differentiated from the conventional reinforcement learning methods by internally modeling the learner, and it enables the instructional policy to optimize even in the setting where the student's learning history is limited. We also evaluated the effectiveness of the proposed method by conducting experiments using a mathematical model. The results show that the proposed framework retained a higher memory rate than existing empirical teaching methods, while suggesting the need to reconsider the learning method of the internal model.
  The current challenges are that it only supports the flash card format and cannot be applied to complex formats such as text-answer formats. Additionally, we were not able to conduct experiments using humans.
  In the future, we plan to test the effectiveness of the proposed method in different settings, such as with different numbers of problems and durations of the experiment. We also aim to experiment with different cognitive models. In addition, we plan to increase the number of instructional methods to be compared and experiment using real students.

\section*{Acknowledgements.}
This research is supported by the  JST-Mirai Program (JPMJMI19B2)  and JSPS KAKENHI (JP19H01129).

\end{document}